\documentclass{article}

\PassOptionsToPackage{numbers, compress}{natbib}
\usepackage[preprint]{neurips_2025}

\usepackage[utf8]{inputenc} 
\usepackage[T1]{fontenc}    
\usepackage{hyperref}       
\usepackage{url}            
\usepackage{booktabs}       
\usepackage{amsfonts}       
\usepackage{nicefrac}       
\usepackage{microtype}      
\usepackage{xcolor}         

\usepackage{graphicx}
\usepackage{amsmath}
\usepackage{xcolor}
\usepackage{xspace}
\usepackage{subcaption}
\usepackage{booktabs}      
\usepackage{pifont}        
\usepackage[table]{xcolor} 

\newcommand{\eg}{\textit{e.g.},\@\xspace}

\newif\ifshowtodos
\showtodostrue     %
\ifshowtodos
  \newcommand{\todo}[1]{\textcolor{red}{[\textit{TODO: #1}]}\xspace}
\else
  \newcommand{\todo}[1]{}
\fi

\title{A Unified 3D Object Perception Framework for Real-Time Outside-In Multi-Camera Systems}

\author{%
  Yizhou~Wang$^{*}$, Sameer~Pusegaonkar$^{*}$, Yuxing~Wang, Anqi~Li, \\
  \textbf{Vishal~Kumar, Chetan~Sethi, Ganapathy~Aiyer, Yun~He, Kartikay~Thakkar,} \\
  \textbf{Swapnil~Rathi, Bhushan~Rupde, Zheng~Tang, Sujit~Biswas} \vspace{1mm}\\
  NVIDIA Corporation\\
  {\small $^*$ Equal contribution}\\
  {\tt\small \{yizwang, spusegaonkar\}@nvidia.com}
}

\begin{document}

\maketitle

\begin{abstract}
  Accurate 3D object perception and multi-target multi-camera (MTMC) tracking are fundamental for the digital transformation of industrial infrastructure. However, transitioning ``inside-out'' autonomous driving models to ``outside-in'' static camera networks presents significant challenges due to heterogeneous camera placements and extreme occlusion. In this paper, we present an adapted Sparse4D framework specifically optimized for large-scale infrastructure environments. Our system leverages absolute world-coordinate geometric priors and introduces an occlusion-aware ReID embedding module to maintain identity stability across distributed sensor networks. To bridge the Sim2Real domain gap without manual labeling, we employ a generative data augmentation strategy using the NVIDIA COSMOS framework, creating diverse environmental styles that enhance the model's appearance-invariance. Evaluated on the AI City Challenge 2025 benchmark, our camera-only framework achieves a state-of-the-art HOTA of 45.22. Furthermore, we address real-time deployment constraints by developing an optimized TensorRT plugin for Multi-Scale Deformable Aggregation (MSDA). Our hardware-accelerated implementation achieves a $2.15\times$ speedup on modern GPU architectures, enabling a single Blackwell-class GPU to support over 64 concurrent camera streams.
\end{abstract}

\section{Introduction}
\label{sec::intro}

Multi-camera perception has become an essential capability in many real-world environments such as warehouses, retail stores, and hospitals. These spaces increasingly rely on networks of static, widely distributed cameras to monitor activity, enhance operational efficiency, and ensure safety. Unlike autonomous driving systems that operate from an ego-centric viewpoint~\cite{li2024bevformer,yang2023bevformer} or leverage synchronized LiDAR--camera fusion~\cite{liang2022bevfusion,huang2022multi,wang2023unitr}, these \textbf{\textit{outside-in}} environments require aggregating information from many heterogeneous static cameras. Large indoor spaces often include significant occlusions from shelving or furniture, disconnected fields of view, lighting variations, and camera configurations that can number from a handful to hundreds. Achieving reliable \textit{multi-target multi-camera tracking} (MTMC) in such conditions remains a challenging problem.

\begin{figure}[t]
\centering
\includegraphics[width=0.7\linewidth]{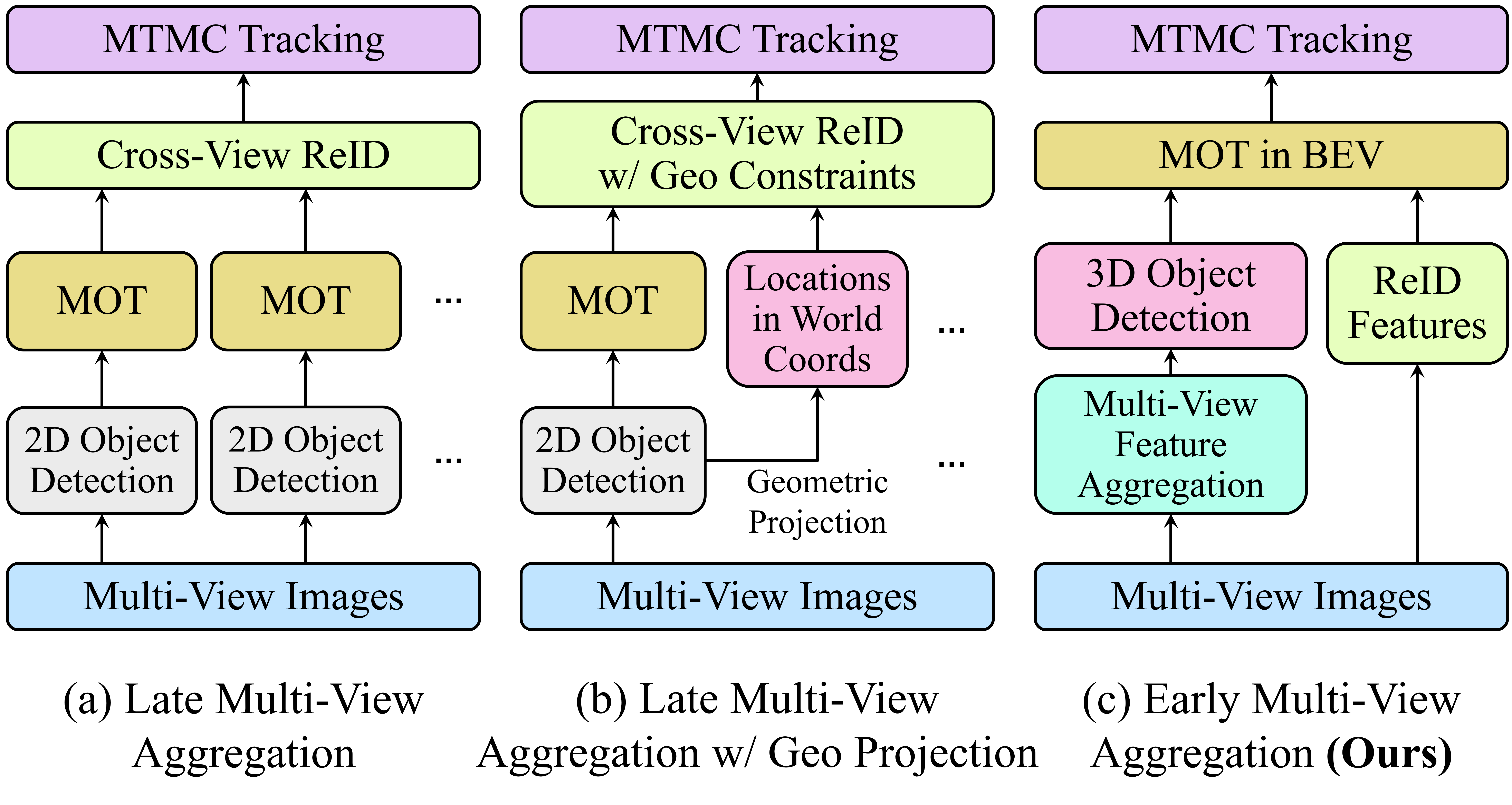}
\caption{Comparison among three types of MTMC tracking methods: (a) conducts 2D detection separately and associates objects among different views by appearance-based ReID; (b) considers geometric constraints as well, besides appearance, for cross-view association; (c) achieves multi-view association in the early stage by feature-level aggregation. Figure is from the MCBLT paper~\cite{Wang_2025_MCBLT}.}
\label{fig:mcblt_types}
\end{figure}

Existing MTMC tracking methods can be broadly grouped into three categories (see Figure~\ref{fig:mcblt_types}), as described in prior work such as EarlyBird\cite{teepe2024earlybird} and MCBLT~\cite{Wang_2025_MCBLT}: 
1) \textbf{Late Aggregation:} Cameras independently perform 2D detection and tracking, followed by cross-camera identity association using appearance features. While modular, this approach is highly sensitive to occlusions and relies heavily on significant camera overlap. 2) \textbf{Projection-Aided Fusion:} These methods project 2D detections into a shared geometric space (e.g., bird's-eye view) to reduce cross-view matching ambiguity. However, performance remains restricted by camera calibration precision and geometric projection errors. 3) \textbf{Early Multi-View Aggregation:} Recent architectures aggregate multi-view features into a unified 3D representation \emph{before} detection, enabling more consistent reasoning. Nevertheless, many existing models struggle with scalability beyond small camera networks or lack robustness in cluttered indoor environments.

At the same time, spatiotemporal perception architectures such as Sparse4D~\cite{lin2022sparse4d,lin2023sparse4dv2,lin2023sparse4dv3} have demonstrated strong performance in ego-centric autonomous driving by integrating multi-sensor features over time using sparse queries. However, directly applying such architectures to static multi-camera environments presents several challenges. Unlike autonomous driving settings, there is no ego-motion to guide temporal alignment, no LiDAR to provide geometric priors, cameras observe widely varying fields of view (FoV), and the system must scale to dozens or even hundreds of cameras while maintaining consistent world-coordinate reasoning.

To address the above issues, this paper proposes a unified outside-in multi-camera 3D object perception framework tailored specifically for MTMC 3D detection and tracking in large infrastructure environments. The system focuses on:
\begin{itemize}
    \item \textbf{Inside-Out to Outside-In Adaptation:} A fundamental re-engineering of the Sparse4D architecture—originally optimized for ego-centric autonomous driving—to handle fixed infrastructure environments. This involves replacing ego-motion-dependent temporal alignment with absolute world-coordinate geometric priors and camera-aware positional encodings to maintain consistency across distributed, static viewpoints.
    \item \textbf{Occlusion-Aware Appearance Embeddings:} A visibility-weighted feature aggregation mechanism that utilizes estimated object velocity and keypoint projection to maintain identity consistency across disjoint and occluded fields of view.
    \item \textbf{Sim2Real Robustness via COSMOS:} A data augmentation pipeline using generative style transfer to bridge the domain gap between synthetic training environments and diverse real-world lighting/texture conditions.
    \item \textbf{Hardware-Accelerated Inference:} A specialized TensorRT plugin for Multi-Scale Deformable Aggregation (MSDA) utilizing FP16 vectorization and asynchronous memory prefetching, achieving up to 2.15$\times$ throughput speedup on modern GPU architectures.
\end{itemize}

Beyond architectural novelty, this work demonstrates state-of-the-art (SOTA) performance on the AI City Challenge 2025 Multi-Camera Tracking leaderboard, specifically within the online, camera-only tracking category. Our system achieves a HOTA of 45.22, significantly outperforming existing online benchmarks. Furthermore, we address the computational demands of large-scale infrastructure. This optimization significantly enhances runtime throughput, enabling the seamless processing of high-density camera networks and ensuring the system's viability for edge-level deployment.


\section{Related Works}
\label{sec::related}

\subsection{Camera-Based 3D Object Perception}

3D object perception using cameras has been extensively explored across multi-camera systems, ego-centric camera rigs, and general multi-view settings. In multi-camera environments, prior work on multi-target multi-camera (MTMC) perception differs primarily in how information across views is fused. Earlier studies, such as EarlyBird~\cite{teepe2024earlybird} and MCBLT~\cite{Wang_2025_MCBLT}, categorize these approaches into late fusion, geometry-aided fusion, and early feature-level fusion.

\paragraph{Late and Geometry-aided fusion.} 
Traditional pipelines process each camera independently, performing 2D detection and within-camera tracking followed by cross-view matching using appearance-based ReID features~\cite{ristani2016performance,zheng2019joint}. While modular, these methods are sensitive to heavy occlusions and limited field-of-view overlap. Subsequent geometry-aided approaches project 2D detections into a shared bird's-eye view (BEV) or coarse 3D space~\cite{tang2019cityflow,luo2021multiple}, often utilizing homography-based projections to ground detections on a reference plane.

\paragraph{Early multi-view feature fusion.} 
Recent advancements focus on early-fusion architectures that aggregate multi-view features into a unified 3D representation before detection. In the autonomous driving domain, dense BEV-based models such as BEVDet~\cite{huang2021bevdet} and BEVFormer~\cite{li2024bevformer,yang2023bevformer} transform image features into a top-down representation for unified reasoning. Simultaneously, sparse query-based methods like DETR3D~\cite{wang2022detr3d}, PETR~\cite{liu2022petr}, and Sparse4D~\cite{lin2022sparse4d} have demonstrated superior efficiency by sampling features directly in 3D space. However, these ``inside-out'' models assume a rigid, synchronized camera rig and rely on ego-motion priors. Our work adapts these concepts to ``outside-in'' infrastructure environments, where camera placements are heterogeneous and lack a common ego-centric reference frame.


\subsection{Temporal Modeling in 3D Object Perception}

Temporal reasoning is essential for maintaining stable 3D object predictions across time. Existing approaches fall into two broad paradigms: \emph{tracking-by-detection} pipelines that operate on 3D detections, and \emph{feature-level temporal fusion} models that integrate temporal information within the perception network.

\paragraph{Two-stage 3D detection and tracking.}
Two-stage pipelines first perform 3D object detection at each frame and then associate detections across time to form trajectories. These methods treat temporal modeling as a data-association problem rather than a feature-level fusion task. MCBLT~\cite{Wang_2025_MCBLT} exemplifies this approach in multi-camera environments by generating sparse 3D detections and applying a graph neural network (GNN) based 3D tracking-by-detection to maintain object identities over long sequences. Similar strategies appear in earlier 3D MOT systems that extend 2D tracking-by-detection into 3D space, such as CenterPoint~\cite{yin2021centerpoint}. While these methods are modular and computationally efficient, they rely heavily on detection quality and do not exploit temporal cues at the feature level.

\paragraph{Feature-level temporal fusion.}
A complementary line of work integrates temporal information directly into the perception network. Query-based spatiotemporal architectures such as Sparse4D~\cite{lin2022sparse4d,lin2023sparse4dv2,lin2023sparse4dv3} and StreamPETR~\cite{wang2023streampetr} propagate object queries across frames using temporal attention, enabling the model to accumulate evidence over time and produce more stable 3D predictions. Other transformer-based temporal fusion approaches track object states implicitly by updating learned token representations over sequences. These methods leverage temporal continuity more effectively but were primarily developed for ego-centric systems, where temporal alignment is facilitated by ego-motion and synchronized sensor data.


\subsection{Generative Synthesis for Sim2Real Robustness}

The performance of 3D perception models in real-world infrastructure is often hampered by the domain gap between synthetic training data and diverse deployment environments. Traditional data augmentation techniques, \eg random cropping and color jittering, are often insufficient to capture the complexity of varying illumination and architectural textures.

Recent works have explored generative AI for high-fidelity style transfer and scene synthesis to bridge this Sim2Real gap. Early unsupervised approaches like CycleGAN~\cite{zhu2017unpaired} demonstrated the potential for domain adaptation without paired data. More recently, 3D-aware generative models such as EG3D~\cite{chan2022efficient} and neural rendering techniques like NeRF~\cite{mildenhall2021nerf} have enabled the synthesis of view-consistent images from sparse observations. In the autonomous driving domain, frameworks such as DriveGAN~\cite{kim2021drivegan} utilize world models to simulate diverse driving scenarios while maintaining temporal and spatial consistency. 

Specifically for infrastructure and large-scale environments, frameworks such as COSMOS~\cite{alhaija2025cosmos} provide unified platforms for generating diverse scene styles---varying lighting, weather, and environmental textures---while strictly preserving the underlying 3D geometric ground truth. Unlike traditional pixel-level style transfer, these generative synthesis methods allow for the creation of large-scale, multi-view datasets that enhance model robustness and generalizability across unseen facilities. By incorporating COSMOS-augmented data, our model learns appearance-invariant features that are critical for long-term identity stability in ``outside-in'' camera networks.

\section{3D Object Perception Model}
\label{sec::model}

Our goal is to perform consistent 3D object perception from outside-in multi-camera systems, where static cameras observe the scene from diverse and potentially non-overlapping viewpoints. Unlike ego-centric multi-camera rigs, these settings lack ego-motion cues and often include heterogeneous camera configurations. Our model unifies multi-view spatial reasoning, occlusion-aware feature aggregation, and temporal consistency into a single framework capable of generating stable 3D object trajectories.

\subsection{Model Architecture}

\begin{figure}[h]
    \centering
    \includegraphics[width=\linewidth]{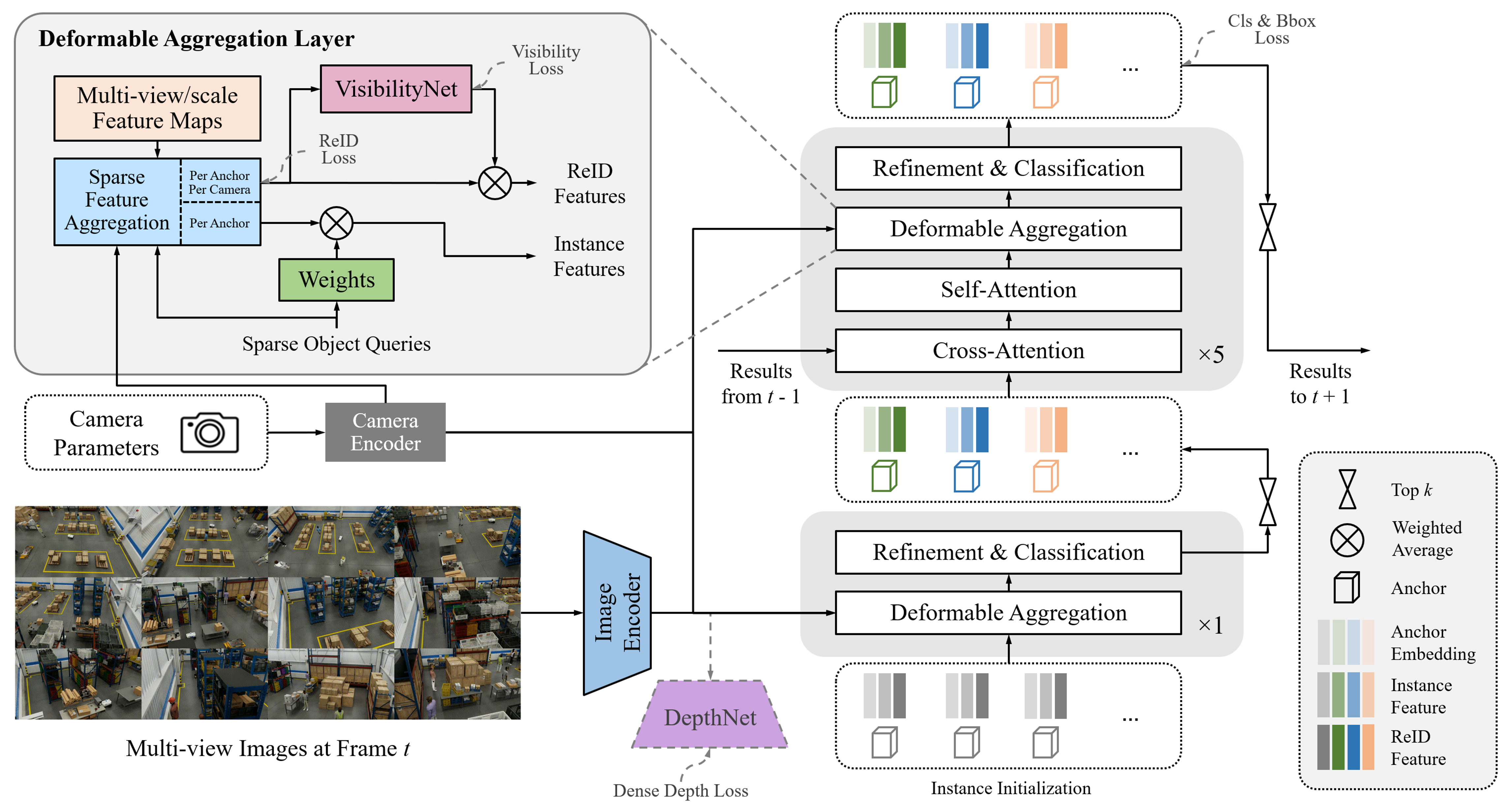}
    \caption{Overview of our 3D object perception model. The system takes multi-view images and camera parameters as input, processes them with an image encoder and camera encoder, and performs multi-view deformable aggregation, occlusion-aware feature fusion, and temporal query propagation to produce consistent 3D object predictions.}
    \label{fig:architecture}
\end{figure}

Figure~\ref{fig:architecture} provides an overview of the model architecture. The system follows a query-based design that integrates multi-view feature extraction, deformable aggregation, and temporal propagation to produce consistent 3D object predictions from outside-in camera networks. Each camera frame is first processed by a shared backbone network to generate multi-scale 2D feature maps, ensuring that all viewpoints contribute features in a consistent representation space.

A set of object queries is maintained across time to represent hypotheses of existing and newly appearing objects. Queries propagated from the previous frame encode both a geometric state and a historical appearance context. Each query includes an estimate of the object’s 3D velocity $(v_x, v_y, v_z)$, which is updated jointly with the object’s spatial parameters. The velocity is predicted by a lightweight regression head attached to the query and refined at each timestep by integrating geometric displacement inferred from multi-view evidence. This velocity estimate plays two roles in the temporal pipeline: it provides a first-order prediction of the object’s future position for initializing queries in the next frame, and it enables motion-compensation in subsequent modules such as occlusion-aware keypoint alignment.

Given the intrinsic and extrinsic parameters of each camera, the 3D anchors associated with each query are projected into the image planes of all cameras. Multi-scale deformable sampling is performed around these projected locations to extract discriminative features from the set of available viewpoints. Temporal consistency is achieved through a recurrent transformer module that fuses the newly gathered multi-view features with the query’s memory feature, enabling the system to accumulate evidence over time and remain stable under appearance changes or partial occlusions. Because outside-in setups lack ego-motion, the model relies on world-coordinate alignment and camera-aware positional encodings instead of ego-motion cues traditionally used in autonomous driving.

After multi-view aggregation and temporal refinement, prediction heads regress the full 3D object state, including spatial parameters $(x, y, z, w, l, h, \text{yaw})$, the 3D velocity vector $(v_x, v_y, v_z)$, the per-object appearance embedding, and a confidence score. The yaw angle provides a compact parameterization of object orientation suitable for the outdoor–indoor scenes and rigid-object categories considered in our system. These jointly predicted quantities reflect both spatial reasoning from multi-view geometry and temporal information accumulated from the recurrent query updates, forming a unified architecture capable of producing accurate and stable 3D object trajectories across heterogeneous camera networks.


\subsection{Occlusion-Aware Embedding} 
\label{sec:occlusion_aware}

\begin{figure}[h]
    \centering
    \includegraphics[width=0.9\linewidth]{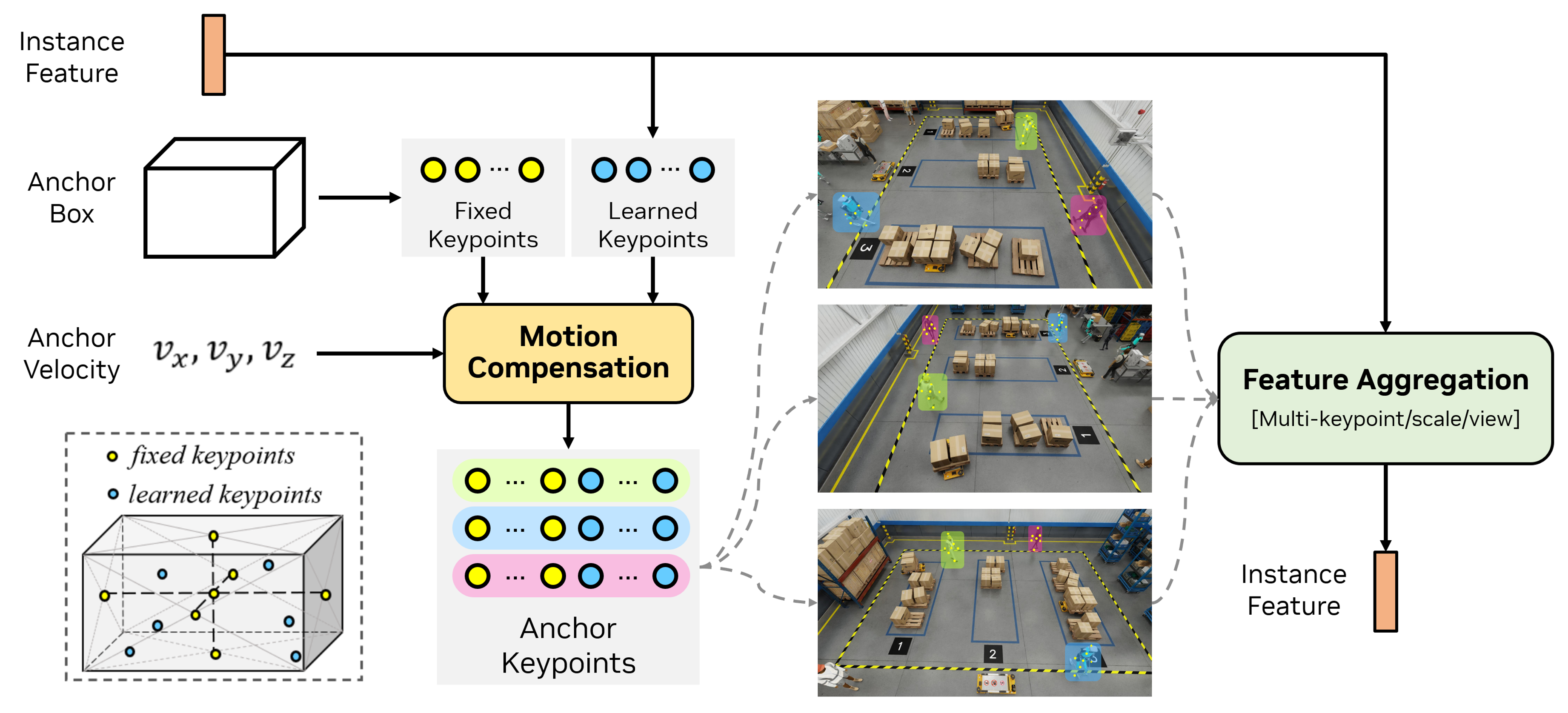}
    \caption{
    Illustration of the occlusion-aware embedding (OAE) module. 
    For each object anchor box, fixed and learned 3D keypoints are projected into each camera view. 
    Motion compensation is applied using the estimated object velocity, producing temporally aligned keypoints. 
    Multi-keypoint, multi-scale, and multi-view image features are aggregated to generate instance-level embeddings. 
    Visibility-aware weighting ensures that occluded or low-quality views contribute less to the final feature representation.}
    \label{fig:occlusion_embedding}
\end{figure}

Outside-in environments exhibit frequent occlusions due to obstacles, varying viewpoints, and heterogeneous coverage. To produce robust per-object embeddings and avoid cross-camera identity fragmentation, we introduce a visibility-weighted fusion mechanism.
Figure~\ref{fig:occlusion_embedding} illustrates the occlusion-aware embedding (OAE) module, which is designed to obtain a robust appearance representation for each object despite heterogeneous viewpoints and frequent occlusions. For every predicted anchor box, we instantiate both fixed geometric keypoints and learned semantic keypoints in 3D space. These keypoints capture complementary cues: the fixed keypoints encode object geometry, while the learned keypoints adapt to regions that consistently provide useful appearance information.

To ensure temporal consistency, the keypoints are aligned using the estimated object velocity $(v_x, v_y, v_z)$. This motion compensation step shifts the keypoints to match the object's expected location in the current frame, resulting in temporally coherent sampling locations that help stabilize the embedding over time.

After motion compensation, each keypoint is projected into each camera view using the known intrinsic and extrinsic parameters. Multi-scale image features at these projected coordinates are extracted from the per-camera backbone features. Since objects may be occluded or partially visible in certain views, we estimate a visibility score $v_{i,k}^t$ for object $k$ in camera $i$ at time $t$. The visibility is defined as:
\begin{equation}
    v_{i,k}^t = \frac{\text{area of visible 2D box}}{\text{area of projected 2D box}}.
\end{equation}
The ground truth for visibility can be obtained directly from the synthetic dataset generated by NVIDIA Omniverse, such as the AI City Challenge 2025 dataset~\cite{Tang_2025_ICCV}. This score reflects both geometric visibility and local feature reliability, and is predicted by a lightweight visibility network.

The final embedding for object $k$ at time $t$ is computed by a visibility-weighted fusion of all per-view features:
\begin{equation}
f_k^t = 
\frac{
\sum_{i=1}^{N} \left\{v_{i,k}^t \cdot g(F_i^t, q_k^t)\right\}
}{
\sum_{i=1}^{N} v_{i,k}^t
},
\end{equation}
where $g(\cdot)$ extracts keypoint-aligned features from the image feature maps $F_i^t$, conditioned on the object query $q_k^t$. The object query $q_k^t$ acts as a high-dimensional feature descriptor that guides the sampling process to relevant spatial regions. This fusion mechanism naturally down-weights unreliable observations while emphasizing informative viewpoints, producing a stable and discriminative representation suitable for identity association in multi-camera environments.


\subsection{Losses}

The model is supervised using geometric, motion, and identity-based losses to produce accurate 3D boxes, reliable yaw estimates, and consistent embeddings. The predicted 3D box parameters $(x, y, z, w, l, h, \text{yaw}, v_x, v_y, v_z)$ are optimized using a smooth regression loss for both spatial and motion components. 
Similar to the original Sparse4D architecture, we incorporate a depth loss ($\mathcal{L}_{\text{depth}}$) to supervise dense depth for improved 3D spatial understanding.
We further introduce a visibility loss ($\mathcal{L}_{\text{vis}}$), implemented as a binary cross-entropy loss, that trains the model to estimate object presence within specific views, enabling the aggregation mechanism to dynamically down-weight occluded features.
Identity consistency is enforced by a cross-entropy loss ($\mathcal{L}_{\text{id}}$) applied to the occlusion-aware embeddings (described in Section~\ref{sec:occlusion_aware}) to promote stable identity propagation across distributed camera views.

The overall objective is defined as a weighted sum:
\begin{equation}
\mathcal{L} = \lambda_{\text{box}}\mathcal{L}_{\text{3D box}} + \lambda_{\text{depth}}\mathcal{L}_{\text{depth}} + \lambda_{\text{vis}}\mathcal{L}_{\text{vis}} + \lambda_{\text{id}}\mathcal{L}_{\text{id}},
\end{equation}
balancing geometric precision, depth-aware localization, and identity coherence in high-density multi-camera settings.

\section{Datasets}
\label{sec::dataset}

We evaluate the proposed system on high-density multi-camera environments that present significant challenges for 3D perception, including varying camera heights, non-overlapping fields of view, and diverse object categories. Our primary benchmark is the AI City Challenge 2025 dataset~\cite{Tang_2025_ICCV}, complemented by a generative Sim2Real augmentation pipeline.

\paragraph{AI City Challenge 2025 Dataset.}
The AI City Challenge 2025 dataset~\cite{Tang_2025_ICCV} serves as the core benchmark for evaluating outside-in multi-camera tracking. This dataset extends the multi-camera configurations of previous editions by introducing a heterogeneous set of moving entities beyond pedestrians. Specifically, the annotations include several categories of industrial service robots and mobile platforms, such as forklifts, NovaCarter robots, Transporter robots, and FourierGR1T2 and AgilityDigit humanoid robots. The inclusion of these diverse categories allows for a rigorous assessment of the model's ability to generalize across varying geometries and motion patterns in complex infrastructure scenes.

For this dataset, we evaluate performance using the Higher Order Tracking Accuracy (HOTA) metric~\cite{luiten2021hota}. HOTA is particularly suited for MTMC tracking as it maintains a balanced decomposition into detection accuracy (DetA), association accuracy (AssA), and localization quality (LocA). This enables a more granular analysis of how our model handles the trade-off between spatial precision and long-term identity maintenance compared to legacy metrics like MOTA or IDF1.

\paragraph{COSMOS Sim2Real Augmentation.}
To improve the robustness of our appearance embeddings against environmental variations, we utilize a generative Sim2Real pipeline based on NVIDIA COSMOS framework~\cite{alhaija2025cosmos}. We created a COSMOS-transferred subset derived from synthetic multi-camera sequences. In this process, original video segments are decomposed into distinct temporal windows, each subjected to a unique text-conditioned style transfer to generate a variety of environmental styles (\eg variations in floor texture, lighting temperature, and atmospheric haze). 

Because the generative process is strictly appearance-based, the underlying 3D geometry remains invariant. This allows us to duplicate the existing ground truth from the synthetic source to the newly stylized videos without the need for manual re-annotation. Training on this augmented dataset enables the model to decouple geometric features from appearance-based cues, significantly enhancing zero-shot generalization when deploying to real-world industrial environments with unseen lighting or texture conditions.

In this study, the Sim2Real pipeline was performed using the NVIDIA Cosmos Transfer 1 Distilled model~\cite{nvidia_cosmos_transfer1_7b_2026}. Each 5-minute video was segmented into 10 clips of 30 seconds in duration. For a camera group comprising 12 distinct viewpoints, this procedure yielded 120 video segments, each of which was processed by Cosmos-Transfer1-Distilled with a distinct simulated text prompt derived from a common template (varying, for example, in time of day, atmosphere, thematic content, environment, weather, and climate). All video segments employed edge control as the sole conditioning signal.

\begin{figure}[t]
    \centering
    \includegraphics[width=0.9\linewidth]{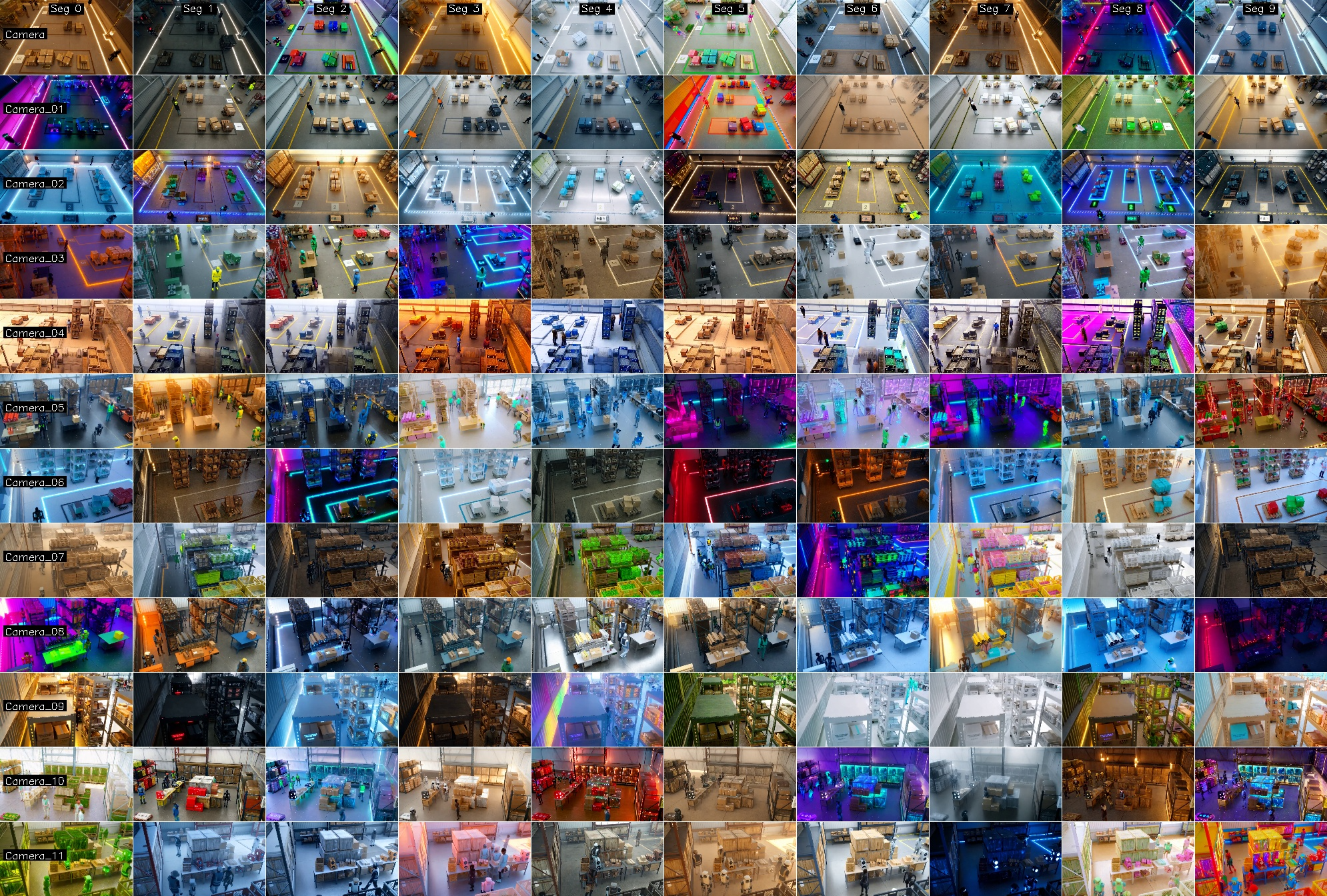}
    \caption{Visualization of the COSMOS-based data augmentation. By applying diverse style transfers to synthetic multi-camera sequences, we generate varied environmental conditions while preserving the exact 3D geometric ground truth.}
    \label{fig:cosmos_dataset}
\end{figure}

\section{Experiments}
\label{sec::experiment}

This section evaluates our framework on the AI City Challenge 2025 benchmark, followed by ablation studies and a performance analysis of our hardware optimizations. We first establish our performance on the official leaderboard, highlighting the efficacy of our camera-only approach for industrial infrastructure. We then assess the individual contributions of our generative Sim2Real training and occlusion-aware ReID module. Finally, we demonstrate the system's real-time throughput across diverse GPU architectures and its zero-shot scalability in unseen environments.

\subsection{Results on AI City Challenge 2025 Dataset}

As summarized in Table~\ref{tab:aicity25_leaderboard}, our Sparse4D framework demonstrates a substantial performance leap over existing online benchmarks. By ascending the leaderboard, we observe that our model surpasses the most comparable online entries by over 13 HOTA points.

A significant performance gap exists between our results and the top-tier methods, such as ZV~\cite{lee2025multi} and SKKU-AutoLab~\cite{tran2025depthtrack}. This discrepancy is primarily attributed to their reliance on auxiliary data, specifically the integration of ground-truth (GT) depth maps alongside RGB imagery. While the inclusion of GT depth significantly enhances 3D detection accuracy and overall HOTA, such data is rarely available in practical environments. In real-world infrastructure deployments, depth sensors are often cost-prohibitive to install at scale, and high-fidelity GT depth maps usually cannot be generated for unconstrained scenes. By relying strictly on camera-based inputs, our framework demonstrates robust generalization without the need for specialized depth hardware, while still outperforming other online competitors by a margin of 13.59 points.

\begin{table}[t]
\centering
\small
\caption{Results on AI City Challenge 2025 dataset on multi-camera tracking. Our Sparse4D model defines the SOTA for online, camera-only perception, significantly outperforming other methods.}
\label{tab:aicity25_leaderboard}
\begin{tabular}{lccc}
\toprule
\textbf{Method} & \textbf{Online} & \textbf{Camera-Only} & \textbf{HOTA $\uparrow$} \\ 
\midrule
UTE AI Lab~\cite{phan2025vgcrtrack} & \checkmark & \ding{55} & 28.03 \\
TeamQDT~\cite{vu2025online} & \checkmark & \ding{55} & 31.63 \\
\midrule
\rowcolor[gray]{0.9} \textbf{Sparse4D (Ours)} & \checkmark & \checkmark & \textbf{45.22} \\
\midrule
SKKU-AutoLab~\cite{tran2025depthtrack} & \ding{55} & \ding{55} & 63.14 \\
ZV~\cite{lee2025multi} & \ding{55} & \ding{55} & 69.91 \\
\bottomrule
\end{tabular}
\end{table}

\paragraph{Sensor Density Scalability.}
Beyond environmental robustness, we further evaluate the framework's architectural flexibility across varying sensor densities. While our primary experiments utilize 32-camera configurations, the system demonstrates remarkable stability when scaled down to sparse, single-camera viewpoints using the same unified checkpoint. A detailed analysis of this zero-shot scalability across different camera counts, along with corresponding visualizations for both sparse and dense networks, is provided in Appendix~\ref{appendix::camera_scalability}.

\subsection{Ablation Studies}


\paragraph{Efficacy of Occlusion-Aware ReID Embeddings.}
Building upon the COSMOS-augmented baseline, we integrate the proposed occlusion-aware ReID embedding module. This final iteration, Sparse4D w/ ReID, achieves our peak HOTA of 45.22, driven by simultaneous gains in detection accuracy and association stability. Notably, the integration of these embeddings also helps recover localization precision compared to the COSMOS-only variant. This recovery indicates that the visibility-weighted fusion mechanism, described in Section~\ref{sec:occlusion_aware}, effectively leverages appearance cues to maintain trajectory continuity even when geometric evidence is weakened by partial occlusions.

\begin{table}[t]
\centering
\small
\caption{Ablation study of Sparse4D on the AI City Challenge 2025 dataset.}
\label{tab:ablation_metrics}
\begin{tabular}{lcccc}
\toprule
\textbf{Model Configuration} & \textbf{HOTA} $\uparrow$ & \textbf{DetA} $\uparrow$ & \textbf{AssA} $\uparrow$ & \textbf{LocA} $\uparrow$ \\ 
\midrule
Sparse4D (Baseline) & 42.18 & 42.74 & 34.89 & \textbf{58.70} \\
Sparse4D w/ COSMOS & 44.71 & 42.69 & 39.01 & 46.50 \\
Sparse4D w/ OAE & \textbf{45.22} & \textbf{43.15} & \textbf{39.43} & 56.57 \\
\bottomrule
\end{tabular}
\end{table}

\begin{figure}[t]
    \centering
    \includegraphics[width=0.5\linewidth]{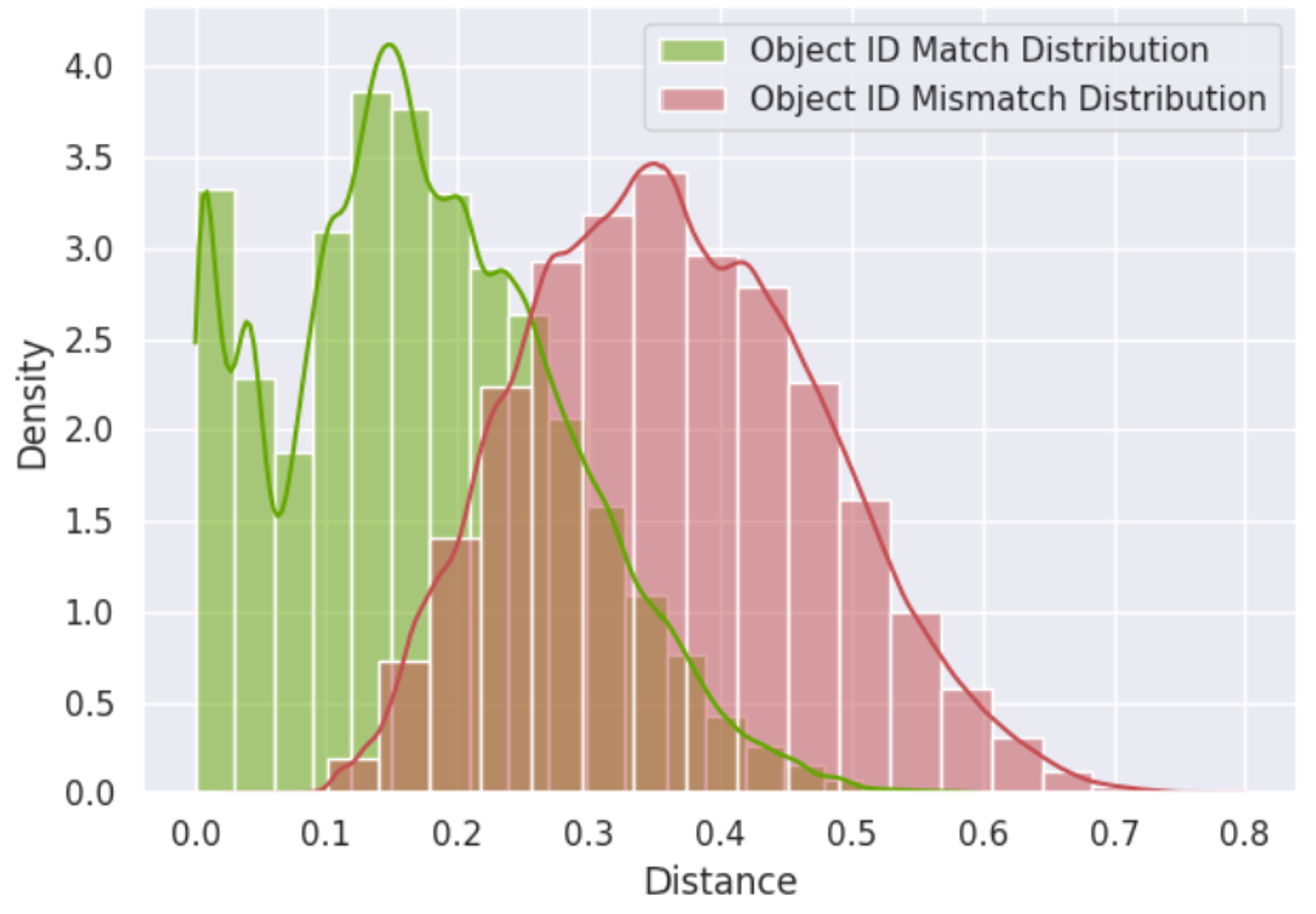}
    \caption{
    Embedding distance distribution for ID matches vs. mismatches  
    }
    \label{fig:embedding_distribution_results}
\end{figure}

\paragraph{Quality of Occlusion-Aware Embedding.}
To quantitatively assess the discriminative power of the learned embeddings, we evaluate the ReID module using a standard retrieval protocol. On this protocol, the module reaches an mAP of 0.7402, with the Cumulative Matching Characteristic (CMC) curve showing a near-perfect Rank-1 retrieval rate of 0.98. As illustrated in Figure~\ref{fig:embedding_distribution_results}, ID matches are heavily concentrated below an L2 distance of 0.3, while mismatches are distributed at significantly higher distances. This clear margin of separation validates the embedding's effectiveness for long-term data association in high-density camera networks.

\subsection{Latency and Throughput Improvements}

To facilitate real-time deployment in high-density camera networks, we developed an optimized FP16-based TensorRT plugin for the Multi-Scale Deformable Aggregation (MSDA) operator, originally introduced in the Sparse4D framework~\cite{lin2022sparse4d}. The optimized kernel leverages half2 vectorization, asynchronous global-to-shared memory copies, and mixed-precision dispatch to maximize throughput while maintaining numerical correctness. Because deformable aggregation involves fusing feature maps from multiple scales and cameras at dynamically predicted locations, the latency of this specific operation is often the primary bottleneck in 3D perception pipelines.

\paragraph{Half2 Vectorization.} 
We implemented half2 vectorization within the MSDA operator to optimize memory bandwidth and arithmetic throughput. By reading and accumulating features as half2 pairs, the system packs two FP16 channels per load, allowing bilinear sampling and interpolation to be performed in parallel. This approach effectively doubles the throughput for arithmetic and memory transactions compared to standard scalar FP16 operations. During execution, the weighted sum remains in half2 format until the final writeback, preserving the packed representation to maximize efficiency.

\paragraph{Asynchronous Prefetching.} 
MSDA operations are typically memory-bound due to low arithmetic intensity and scattered memory access patterns as shown in ~\cite{lin2022sparse4d}. To mitigate this, we utilize asynchronous prefetching to overlap data movement with computation. The kernel pulls feature map tiles into shared memory while the Streaming Multiprocessor (SM) processes the previously fetched tile. This decoupling of load and compute phases reduces exposed DRAM latency and increases SM utilization, effectively mitigating the memory stalls that traditionally bottleneck deformable aggregation.

\begin{table}[t]
\centering
\small
\caption{Comparison of perception metrics between the baseline MSDA kernel and the optimized FP16 implementation. The results demonstrate that the optimization maintains numerical parity.}
\label{tab:kernel_parity}
\begin{tabular}{lcccc}
\toprule
 & HOTA & DetA & AssA & LocA \\
\midrule
Original MSDA Kernel & 41.106 & 43.149 & 39.429 & 56.573 \\
Optimized MSDA Kernel & 41.105 & 43.147 & 39.428 & 56.573 \\
\bottomrule
\end{tabular}
\end{table}

\paragraph{Hardware Throughput Comparison.}
We evaluated the throughput of the optimized Sparse4D framework across a comprehensive suite of NVIDIA hardware. As shown in Table~\ref{tab:gpu_speedup}, the optimized kernel provides significant speedups across all architectures, with the Blackwell-based B200 and GB200 achieving gains of over \textbf{2.15$\times$}. The full comparison table, detailing performance metrics across distinct GPU configurations, including edge and workstation modules, is provided in Appendix~\ref{appendix::throughput}.

\begin{table}[t]
\centering
\small
\caption{Throughput comparison on the number of cameras supported at 30 FPS for Sparse4D with a ResNet-101 backbone.}
\label{tab:gpu_speedup}
\begin{tabular}{lccc}
\toprule
GPU Variant & Baseline Sparse4D & Accelerated MSDA & Speedup \\
\midrule
A100 & 8 & 12 & $1.50\times$ \\
H100 & 14 & 18 & $1.29\times$ \\
B200 & 26 & 56 & $2.15\times$ \\
GB200 & 31 & 64 & $2.06\times$ \\
Jetson AGX Thor & 2 & 4 & $2.00\times$ \\
DGX Spark & 2 & 3 & $1.50\times$ \\
\bottomrule
\end{tabular}
\end{table}

\subsection{Model Generalizability}

To evaluate the robustness of our framework beyond the primary benchmark, we investigate its performance through generative Sim2Real training and zero-shot deployment in unseen environments.

\paragraph{Generative Sim2Real Augmentation using COSMOS.} 
To bridge the domain gap between synthetic training environments and real-world industrial textures, we employ a generative augmentation strategy based on the COSMOS framework. We utilize a hybrid training split of 80:20, where 80\% of the samples are derived from the base synthetic dataset and the remaining 20\% are COSMOS-stylized sequences. This approach allows the model to encounter a vast diversity of lighting conditions and floor textures while preserving the precise geometric grounding of the original synthetic labels. As shown in Table~\ref{tab:ablation_metrics}, training with this generative injection resulted in a HOTA increase from 42.18 to 44.71 before the final addition of ReID embeddings, confirming that text-conditioned style transfer bolsters the model's appearance-invariance.

\paragraph{Zero-Shot Performance in Diverse Environments.}
The generalizability of our system is further validated through zero-shot inference on large-scale, unseen datasets, specifically the \textbf{Building-K} and \textbf{Large-Scale Synthetic Warehouse} benchmarks. Without additional fine-tuning, our model demonstrates high spatial stability and tracking continuity in these novel environments. Please see the qualitative zero-shot results in Appendix~\ref{appendix::zero-shot}.

\section{Conclusion}
\label{sec::conclusion}

In this work, we have presented an adaptation of the Sparse4D architecture specifically engineered for "outside-in" infrastructure-based 3D perception. By re-engineering the model to leverage absolute world-coordinate geometric priors instead of ego-motion-dependent alignment, we successfully transitioned a leading autonomous driving framework into the domain of static camera networks.

Our experimental results on the AI City Challenge 2025 leaderboard demonstrate that our final iteration, which integrates generative Sim2Real augmentation via the COSMOS framework and specialized occlusion-aware ReID embeddings, achieves a state-of-the-art HOTA of 45.22. This represents a significant margin over existing online competitors and establishes a new benchmark for camera-only 3D tracking in industrial environments.

Furthermore, we addressed the computational challenges of high-density camera deployments by introducing an optimized TensorRT plugin for Multi-Scale Deformable Aggregation. Our hardware-accelerated implementation achieved a throughput increase of over 2.15$\times$ on modern GPU architectures, enabling the real-time processing of over 64 concurrent camera streams on a single Blackwell-class GPU.

\clearpage
\section*{Appendices}
\appendix
\section{Zero-Shot Inference Results}
\label{appendix::zero-shot}

To evaluate the cross-domain robustness of the Sparse4D framework, we conducted zero-shot inference across distinct synthetic and real-world industrial environments. These tests were performed without environment-specific fine-tuning, relying solely on the generalized features learned during the generative COSMOS-augmented training phase.

\begin{figure}[h]
    \centering
    \includegraphics[width=0.95\linewidth]{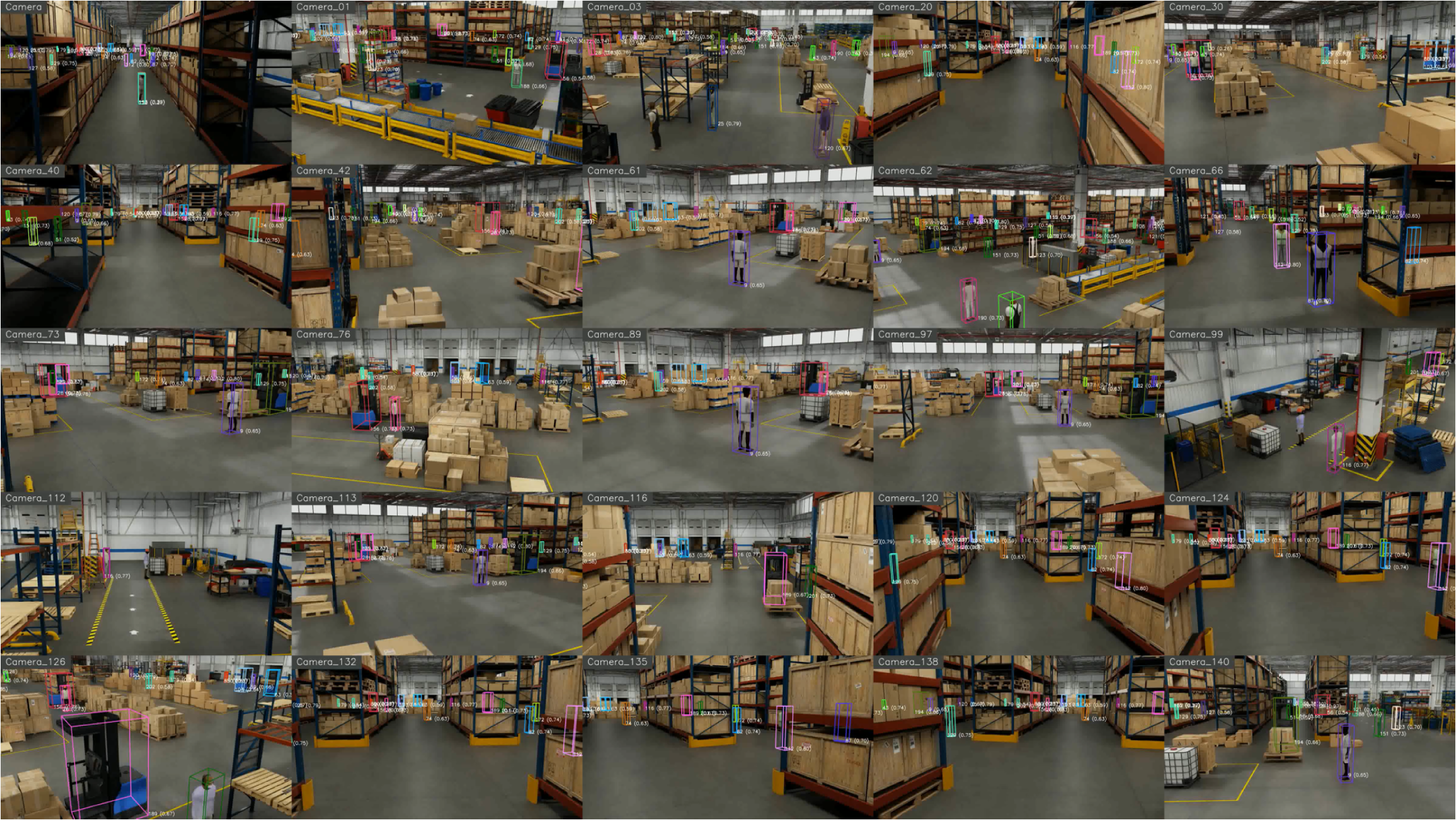}
    \caption{Visualization of zero-shot inference on the large-scale \textit{synthetic} warehouse dataset (25 cameras). The framework maintains global trajectory continuity across a dense, overlapping sensor network.}
    \label{fig:large_scale_synthetic_warehouse_zero_shot_inference}
\end{figure}

As shown in Figure~\ref{fig:large_scale_synthetic_warehouse_zero_shot_inference}, the framework successfully maintains identity stability and precise 3D localization across the diverse camera configurations and high object densities of a large-scale synthetic warehouse. This 25-camera setup demonstrates the system's ability to handle significant viewpoint overlap and long-range tracking in structured environments.

\begin{figure}[h]
    \centering
    \includegraphics[width=0.95\linewidth]{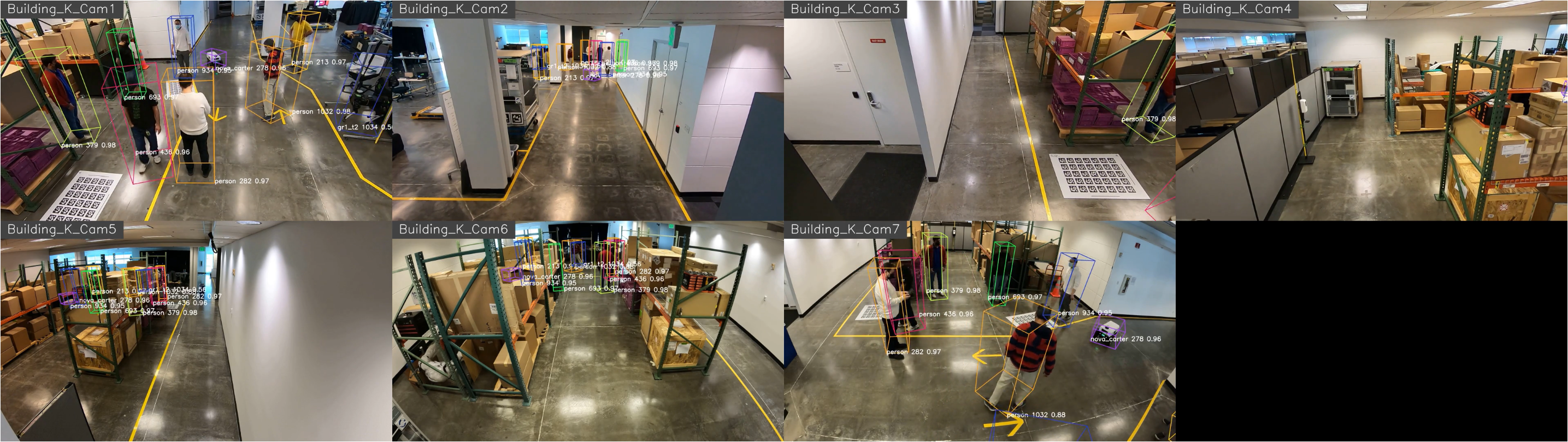}
    \caption{Visualization of zero-shot inference on the \textit{real-world} Building-K dataset. The results demonstrate the framework's robustness to domain shifts in lighting and texture when deployed in actual infrastructure.}
    \label{fig:buildingk_zero_shot_inference}
\end{figure}

Furthermore, the model's generalizability is validated on the real-world Building-K dataset, as illustrated in Figure~\ref{fig:buildingk_zero_shot_inference}. Despite the shift from synthetic textures to unconstrained real-world lighting and physical sensor noise, the framework maintains stable trajectories for pedestrians and moving equipment. This performance confirms that the integration of absolute world-coordinate geometric priors and occlusion-aware embeddings allows for reliable ``outside-in'' perception in novel infrastructure deployments without the need for additional manual labeling or site-specific adaptation.

\section{Scalability Across Variable Camera Densities}
\label{appendix::camera_scalability}

A core strength of our adapted Sparse4D framework is its architectural flexibility, allowing it to operate across a wide range of sensor densities without requiring model retraining or architecture modifications. The same unified checkpoint demonstrates robust performance when deployed on configurations ranging from a single-camera viewpoint to a dense 32-camera network. 

\begin{figure}[h]
    \centering
    \includegraphics[width=0.6\linewidth]{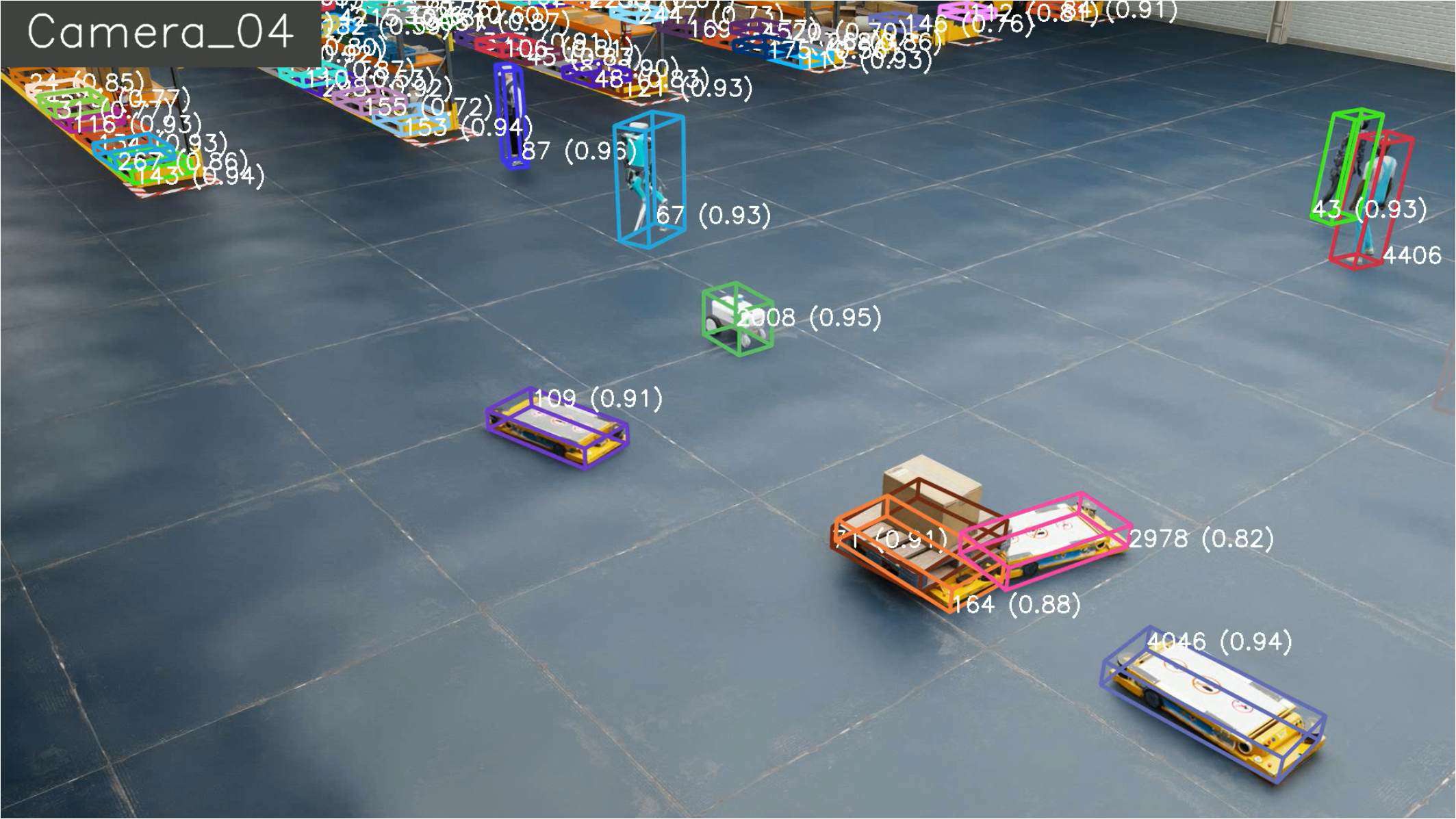}
    \caption{Visualization of Sparse4D results for single-camera 3D detection and tracking, demonstrating performance stability in sparse sensor environments.}
    \label{fig:results_single_view}
\end{figure}

The system's adaptability to sparse sensor configurations is illustrated in Figure~\ref{fig:results_single_view}, where the model maintains precise 3D localization and identity continuity using only a single-camera perspective.

While theoretical architectural limits suggest that a single Sparse4D model can manage over 60 concurrent camera inputs, practical dataset constraints limited our experimental validation to high-density scenes with up to 32 cameras. In these tests, the system exhibited consistent performance, effectively handling the high occlusion rates and overlapping fields of view typical of distributed infrastructure sensing. 

\begin{figure}[h]
    \centering
    \includegraphics[width=\linewidth]{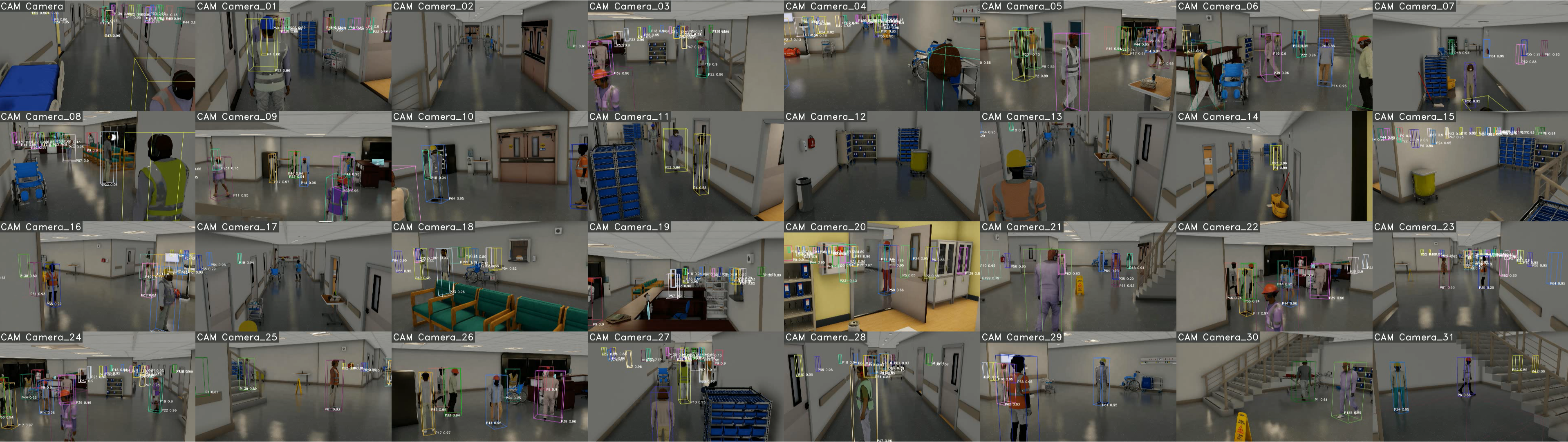}
    \caption{Visualization of Sparse4D results for a 32-camera synthetic hospital environment. The framework maintains global identity stability and 3D localization across a high-density camera network using a single unified checkpoint.}
    \label{fig:results_hospital}
\end{figure}

The capacity of the framework to scale to massive, distributed sensor networks is demonstrated in Figure~\ref{fig:results_hospital}, which visualizes the zero-shot inference results on a high-density, 32-camera synthetic hospital scene. Despite the complex spatial layout and significant camera overlap, the unified query-based architecture ensures stable tracking across the entire facility.

\section{Detailed Throughput Experiments}
\label{appendix::throughput}

\begin{figure}[h]
    \centering
    \includegraphics[width=\linewidth]{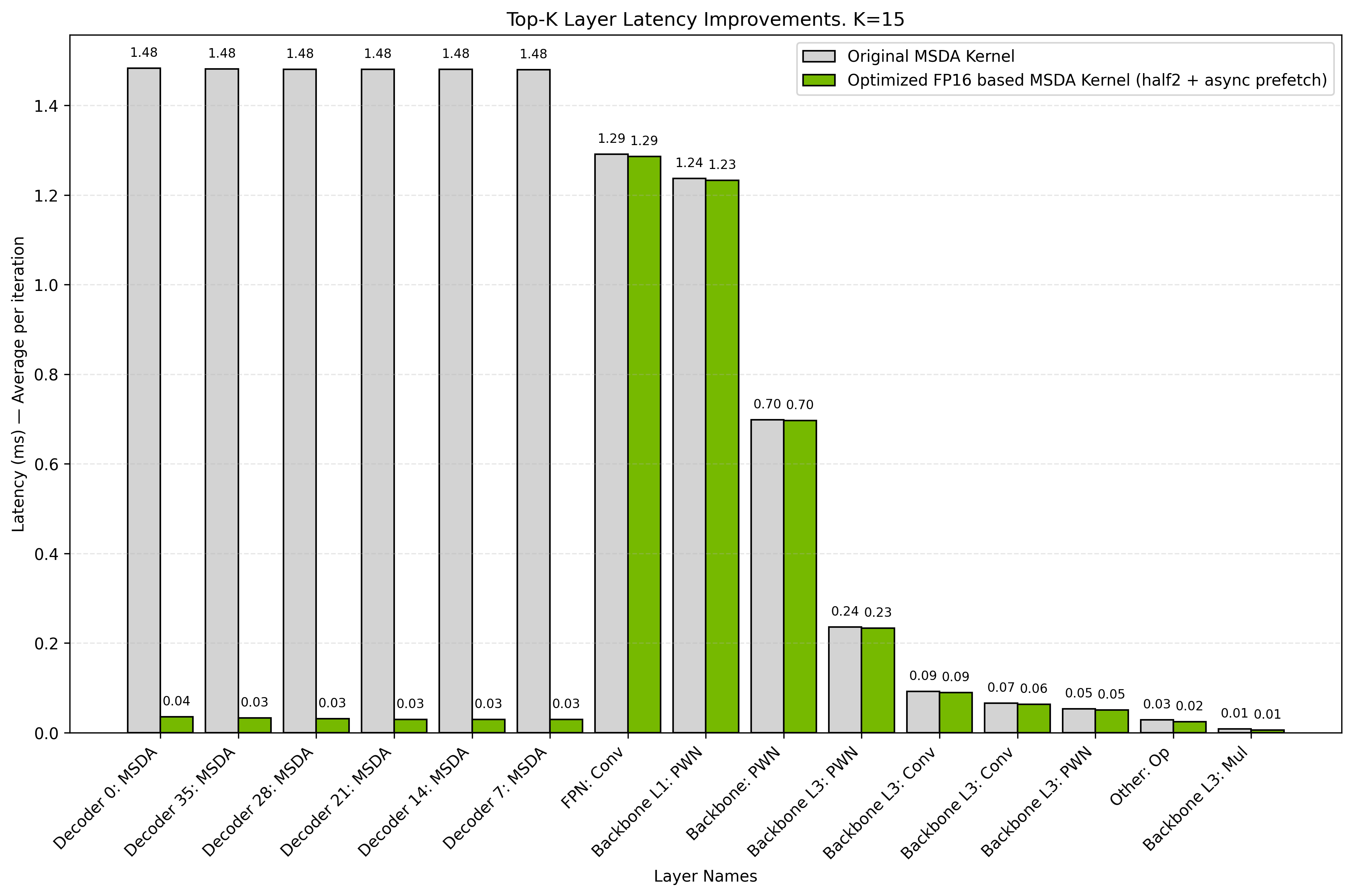}
    \caption{
    Layer-wise latency comparison of the top-15 layers, showing differences between the original MSDA kernel and the optimized kernel. Measurement is obtained on an H100 GPU.
    }
    \label{fig:msda_fp16}
\end{figure}

\paragraph{Layer-wise Latency Breakdown of MSDA Optimization.}
Figure~\ref{fig:msda_fp16} presents a detailed layer-wise latency breakdown for the top-15 most computationally intensive layers of the model, as measured on an NVIDIA H100 GPU. The bar chart provides a direct comparison between the average latency per iteration of the original Multi-Scale Deformable Aggregation (MSDA) kernel (depicted in gray) and the optimized FP16-based MSDA kernel, which incorporates half2 vectorization and asynchronous prefetching (depicted in green). The results demonstrate a substantial performance improvement in the MSDA-specific layers, which were previously the primary computational bottlenecks. For example, the latency for ``Decoder 0: MSDA'' drops significantly from 1.48 ms with the original kernel to just 0.04 ms with the optimized version. Similar substantial reductions are observed across all other Decoder MSDA layers shown. Conversely, non-MSDA layers such as ``FPN: Conv'' and various backbone layers exhibit negligible latency differences, confirming that the optimizations effectively target the intended operations without negatively impacting other parts of the network.

\paragraph{Full Throughput Comparison.} As detailed in Table~\ref{tab:gpu_speedup_full}, the optimized MSDA kernel provides consistent speedups across the entire hardware spectrum. The most significant gains are observed on Blackwell-class architectures, where the hardware's native support for asynchronous memory operations allows the framework to scale up to 64 concurrent camera streams at 30 FPS.

\begin{table}[h]
\centering
\small
\caption{Throughput comparison on the number of cameras supported at 30 FPS for Sparse4D with a ResNet-101 backbone.}
\label{tab:gpu_speedup_full}
\begin{tabular}{lccc}
\toprule
GPU Variant & Baseline Sparse4D & Accelerated MSDA & Speedup \\
\midrule
A100 & 8 & 12 & $1.50\times$ \\
H100 & 14 & 18 & $1.29\times$ \\
H200 & 15 & 18 & $1.20\times$ \\
B200 & 26 & 56 & $2.15\times$ \\
GB200 & 31 & 64 & $2.06\times$ \\
L4 & 2 & 3 & $1.50\times$ \\
L40S & 8 & 10 & $1.25\times$ \\
RTX PRO 6000 Server Edition & 12 & 18 & $1.50\times$ \\
RTX PRO 6000 Workstation Edition & 12 & 20 & $1.67\times$ \\
RTX 6000 Ada & 8 & 9 & $1.13\times$ \\
Jetson AGX Thor -- T5000 & 2 & 4 & $2.00\times$ \\
DGX Spark & 2 & 3 & $1.50\times$ \\
\bottomrule
\end{tabular}
\end{table}

\paragraph{Real-time Scalability.}
The system's real-time scalability was further analyzed using the NVIDIA DeepStream pipeline~\cite{nvidia_deepstream}. Table~\ref{tab:deepstream_throughput} summarizes the number of concurrent camera streams supported at various frame rates. The TensorRT metrics reflect model-only latency, while the DeepStream metrics include additional overhead from instance-bank pre- and post-processing.

\begin{table}[h]
\centering
\caption{Real-time camera support capacity for Sparse4D. DeepStream columns include system-level pre- and post-processing overhead.}
\label{tab:deepstream_throughput}
\resizebox{\linewidth}{!}{%
\begin{tabular}{lcccccc}
\toprule
GPU Variant &
\shortstack{TensorRT\\@30 FPS} &
\shortstack{DeepStream\\@30 FPS} &
\shortstack{TensorRT\\@15 FPS} &
\shortstack{DeepStream\\@15 FPS} &
\shortstack{TensorRT\\@10 FPS} &
\shortstack{DeepStream\\@10 FPS} \\
\midrule
DGX Spark & 3 & 2 & 7 & 5 & 11 & 8 \\
RTX PRO 6000 Server Edition & 18 & 13 & 38 & 29 & 59 & 45 \\
RTX PRO 6000 Workstation Edition & 20 & 15 & 40 & 30 & 60 & 46 \\
Jetson AGX Thor -- T5000 & 4 & 3 & 9 & 6 & 14 & 10 \\
B200 & $>64$ & 43 & $>64$ & $>49$ & $>64$ & $>49$ \\
GB200 & 64 & 49 & $>64$ & $>49$ & $>64$ & $>49$ \\
H100 SXM HBM3 -- 80GB & 18 & 13 & 22 & 16 & 35 & 26 \\
H200 & 18 & 13 & 29 & 22 & 46 & 35 \\
RTX 6000 ADA & 9 & 6 & 18 & 13 & 27 & 20 \\
A100-SXM4 -- 80GB & 12 & 9 & 27 & 20 & 42 & 32 \\
L4 -- 24GB & 3 & 2 & 7 & 5 & 10 & 7 \\
L40S -- 48GB & 10 & 7 & 20 & 15 & 31 & 23 \\
\bottomrule
\end{tabular}%
}
\end{table}

\clearpage
\setcitestyle{numbers}
\bibliographystyle{plainnat}
\bibliography{main}

\end{document}